\documentclass[conference, letterpaper]{IEEEtran}
\IEEEoverridecommandlockouts

\usepackage{cite}
\usepackage{amsmath,amssymb,amsfonts}

\usepackage{graphicx}
\usepackage{textcomp}
\usepackage{xcolor}
\usepackage{algorithm} 
\usepackage{algpseudocode}
\usepackage{caption}
\def\BibTeX{{\rm B\kern-.05em{\sc i\kern-.025em b}\kern-.08em
    T\kern-.1667em\lower.7ex\hbox{E}\kern-.125emX}}
\begin{document}

\makeatletter 
\newcommand{\linebreakand}{%
  \end{@IEEEauthorhalign}
  \hfill\mbox{}\par
  \mbox{}\hfill\begin{@IEEEauthorhalign}
}
\makeatother

\title{Metalearning traffic assignment for network disruptions with graph convolutional neural networks*\\
\thanks{Novo Nordisk Foundation grant NNF23OC0085356.}
}

\author{\IEEEauthorblockN{1\textsuperscript{st} Agriesti Serio}
\IEEEauthorblockA{\textit{Department of Technology, Management and Economics,} \\
\textit{Technical University of Denmark,}\\
Lyngby, Denmark \\
samaa@dtu.dk}
\and
\IEEEauthorblockN{2\textsuperscript{nd} Cantelmo Guido}
\IEEEauthorblockA{\textit{Department of Technology, Management and Economics,} \\
\textit{Technical University of Denmark,}\\
Lyngby, Denmark \\
guica@dtu.dk}
\linebreakand
\IEEEauthorblockN{3\textsuperscript{rd} Pereira Francisco Camara}
\IEEEauthorblockA{\textit{Department of Technology, Management and Economics,} \\
\textit{Technical University of Denmark,}\\
Lyngby, Denmark \\
camara@dtu.dk}}

\maketitle

\begin{abstract}
Building machine-learning models for estimating traffic flows from OD matrices requires an appropriate design of the training process and a training dataset spanning over multiple regimes and dynamics. As machine-learning models rely heavily on historical data, their predictions are typically accurate only when future traffic patterns resemble those observed during training. However, their performance often degrades when there is a significant statistical discrepancy between historical and future conditions. This issue is particularly relevant in traffic forecasting when predictions are required for modified versions of the network, where the underlying graph structure changes due to events such as maintenance, public demonstrations, flooding, or other extreme disruptions. Ironically, these are precisely the situations in which reliable traffic predictions are most needed. In this case, the number of relevant features, values and graphs expands very quickly with the number of links and zones in the case study. This makes it difficult, especially for transport practitioners, to design training datasets that will conceivably cover all the possible distribution shifts arising from changes in demand patterns and/or network disruptions. In the presented work, we combine a machine-learning model (graph convolutional neural network) with a meta-learning architecture to train the former to quickly adapt to new graph structures and demand patterns, so that it may easily be applied to scenarios in which changes in the road network (the graph) and the demand (the node features) happen simultaneously. Our results show that the use of meta-learning allows the graph neural network to quickly adapt to unseen graphs (network closures) and OD matrixes while easing the burden of designing a training dataset that covers all relevant patterns for the practitioners. The proposed architecture achieves a $R^2$ of around 0.85 over unseen closures and OD matrixes.
\end{abstract}


\section{Introduction} \label{sec:intro}
Transportation problems are often visualized as graph structures due to the nature of the data that is used in the field. A population of travelling individuals, for example, can be easily be framed as a social network made of nodes and connections. The traffic that they generate, in turn, spreads across a network composed by links and junctions, namely a graph. 
Thus, alignment between traffic problems and Graph Neural Networks (GNNs) has been widely discussed in literature \cite{b1,b2} and applied to intelligent transportation systems \cite{b3}, to the forecasting of crash \cite{b4}, speed \cite{b5}, flow \cite{b6,b7} and density \cite{b8}, to automated mobility on-demand control problems \cite{b9} and to cross-city knowledge transfer \cite{b10}. 

Relevant to the presented work is literature that focuses on machine-learning models for traffic assignment that results in link flows across the network, function of different OD matrixes. In this work, we refer to machine-learning models as surrogate models, understood as any tool learning to map input-output relationships. 
In \cite{b1}, the authors use the GraphGPS framework \cite{b11} to build a surrogate model for the Traffic Assignment Problem (TAP) with a fixed network. Similarly, in \cite{b2}, traffic flows are learned from ODs through a Graph Convolution Neural Network and a diffusion mechanism. The authors of \cite{b12} also build a surrogate for the traffic assignment through a GNN, as they build the graph by adding virtual links between key locations to ensure a message passing reflecting the correct topological representation. In \cite{b13}, an attention mechanism is embedded in the GNN architecture to build a surrogate for the assignment of different vehicle classes. The work of \cite{b14} results instead in a graph attention mechanism tested on multiple scenarios with different OD matrices and capacities. 

Overall, many works in literature study how a certain architecture can learn the traffic assignment of flows on a specific graph. One key limitation though is that the graph itself, with few exceptions, is rather fixed while the boundary conditions are changed to test the robustness of the proposed models. Namely, each architecture solves the traffic assignment on a specific network (or set of networks) but the problem of a network disrupted in its graph structure is less tackled in literature. In this work, we embed a GNN model in a meta-learning framework (MAML \cite{bmeta}) with the explicit aim of training it to achieve the best starting point for fast adaptation to different graph structures over the road network. 
\begin{figure*}[!t]
    \centering
    \includegraphics[width=0.8\linewidth]{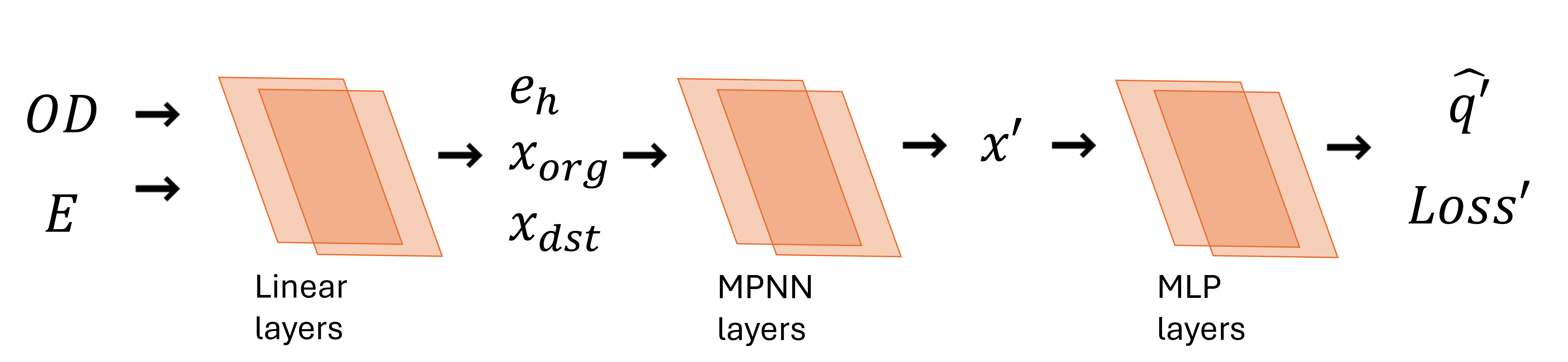}
    \caption{Schematic representation of one training epoch for the gatedGCN. $OD$ and $E$ are matrixes with information about the OD matrix and the edges in the graph. $\hat{q}$ represents instead the estimated flows and $Loss' = SmoothL1Loss$ \cite{b23}}
    \label{fig:gatedgcn}
\end{figure*}
\subsection{Meta-learning in transportation}
Meta-learning is a methodological approach that aims to train a machine-learning model so that it learns how to solve tasks. In practice, the meta-learning framework trains weights that represent the best starting point for quick initialization in tasks that share a similar structure with the ones seen during training. A task can be a regression from input to output, a classification or a reinforcement learning problem, for example \cite{bmeta}. In transportation, meta-learning has been successfully applied to the estimation of the macroscopic fundamental diagram in data-scarcity conditions \cite{b16}, as the authors trained a multitask physics-informed neural network on cities with a lot of data and applied the initialization to cities with less data for the estimation. Another work in the field leveraging meta-learning is \cite{b17}, in which the authors exploit meta-learning to predict multi-modal trajectories by transferring representations from modes with more trajectory records to modes with fewer trajectories. In \cite{b18}, the prediction of multi-modal demand is carried out through a spatio-temporal meta-parameter learning method, while in \cite{b19} meta-learning is used to estimate travel time per mode of transport. 
Thus, as a framework, meta-learning has been applied to multiple transport problems.

\subsection{Meta-learning for traffic assignment}
In our work, we exploit meta-learning with each traffic assignment on a specific network as a task. The objective is for the meta-learning framework to initialize the GNN weights in a way that allows for fast adaptation on an unseen graph. From a transport point of view, we build a MAML+GNN architecture that can be used on the cases in which the graph changes in structure, due for example to network maintenance \cite{b20}, large-scale events such as public demonstrations or disruptions such as flooding \cite{b21} and other extreme weather events. All of these changes in the network structure also cause changes in the OD demands, which increases the space of features the GNN has to map beyond historical patterns. In Section~\ref{sec:methodology} we describe how the selected GNN - a Graph Convolutional Neural Network (GCN) - has been embedded in a variation of the MAML framework introduced in \cite{bmeta}; In Section~\ref{sec:exp-set} we report our data generation process and our experimental settings; In Section~\ref{sec:results} we report our results and comment on them, highlighting merits and limitations of the work; In Section~\ref{sec:conclusions} we discuss the conclusions.

\section{Methodology}\label{sec:methodology}
The proposed framework embeds a gatedGCN model in a MAML architecture, with the aim of initializing the GCN weights for quick adaptation to new graphs and OD matrices. The two main building blocks (the gatedGCN model and MAML) are described in the following: 

\subsection{The gatedGCN}
The surrogate is tailored over the task of learning one single assignment over one graph, with multiple ODs. The selected features are also representative of the nature of the problem, with node features being: the number of edges converging, diverging from the node, the sum of the two and a flattened vector with the flow generated towards all the other destinations. The edge features, on the other hand, include capacity and a "present" flag that will be used to model different networks once the model is embedded in the MAML. Edges with "present" flagged as zeros will be hidden by a mask, severing the connection. The listed features are lifted to the latent space using two linear layers (one for nodes and one for edges). These hidden representations are fed to the main component of the gatedGCN, the message passing layers (MPNN), each mathematically characterized by the following equations:

\begin{equation}
    g = \sigma(W_ee_h+W_{dst}x_{dst}+W_{org}x_{org})\\ 
\end{equation}
\begin{equation}
    m = (W_mx_{org})\odot g\\
\end{equation}
\begin{equation}
    agg_{dst}=\frac{\sum{m}}{\sum{g+\epsilon}} \\
\end{equation}
\begin{equation}
    x' = W_{self}x_h+agg
\end{equation}

In the reported mathematical formulation $W_e, W_{org}, W_{dst}, W_m, W_{self}$ are the weights the neural network applies respectively to the edge features $e_h$, to the features $x_{org}, x_{dst}$ of respectively the origin and destination nodes, to the message $m$ that is passed between nodes and the hidden features $x_{org}$ of each node. $\epsilon$ is a small value to avoid dividing by zero and $x'$ is the updated representation of each node $x_h$ after each MPNN layer. The output of the MPNN layers is converted into edge flows through a 3 layers MLP. A high level schematic representation of each training epoch is provided in Fig.~\ref{fig:gatedgcn}. The used gatedGCN architecture aligns with a base implementation from literature. For a general overview on gated GCNs, the reader is referred to \cite{b22}. 

\subsection{The MAML architecture}

Once the gatedGCN has been designed, it can be embedded in the wider architecture that is MAML. As mentioned in Section~\ref{sec:intro}, meta-learning aims at initializing the weights of a surrogate model (in this case the gatedGCN) so that it quickly initializes to new experiments, without the need to retrain the surrogate with large amount of new data. To do so, MAML simulates a situation of data scarcity in its own training regime and evaluates how well the surrogate model fares, validating the various predictions on the full set of data. It does so by training the surrogate in two phases: inner-loop and outer-loop. During the inner-loop, the surrogate is shown little data concerning a task and is tasked to learn weights that minimize its own loss function wrt. said data. Once this is done, in the outer-loop the surrogate is shown the full amount data and its performance is evaluated against this larger benchmark. In our case, we want the gatedGCN to make good flow predictions for new network closures that have little data available (few OD matrix assignments). Thus, the MAML architecture should train the gatedGCN over different tasks (network closures) first showing it few OD matrix assignments (namely what is called, in MAML literature, support set) and then validating the gatedGCN predictions over more OD matrix assignments (query set). The algorithm is reported in Alg.\ref{alg:maml}.

\begin{algorithm}[H]
\caption{Meta-training an adapted MAML for Few-Shot Supervised Learning}\label{alg:maml}
\begin{algorithmic}
    \Require{$p(\mathcal{T})$: distribution over tasks}
    \Require{$\alpha, \beta$: step size hyperparameters (learning rates)}
    \State randomly initialize $\theta$
    \While{not done}
        \State Initialize outer loop
        \State Sample batch of tasks $\{\mathcal{T}_i\} \sim p(\mathcal{T})$
        \For{\textbf{all} $\mathcal{T}_i$}
            \State Sample $K$ obs. $\mathcal{D}^\text{support}_i = \{x^{(j)}, y^{(j)}\}_{j=1}^K$ from $\mathcal{T}_i$
            \State Sample $M$ obs. $\mathcal{D}^\text{query}_i = \{x^{(j)}, y^{(j)}\}_{j=1}^M$ from $\mathcal{T}_i$
            \State Initialize task-specific parameters $\theta'_i \gets \theta$
            \For{\textbf{$N_{ite}$ inner loop steps}}
                \State Compute K-specific loss:
                \State $\mathcal{L}_\text{K}(\mathcal{D}^\text{support}_i; \theta'_i)$
                \State Average loss across $K$ samples:
                \State $\mathcal{L}_\text{Task}=\frac{1}{K}\sum \mathcal{L}_{K}$
                \State Update task-specific parameters:
                \State $\theta'_i \gets \theta'_i - \alpha \nabla_{\theta'_i} \mathcal{L}_\text{task}(\mathcal{D}^\text{support}_i; \theta'_i)$
            \EndFor
            \State Compute query loss: $\mathcal{L}_\text{task}(\mathcal{D}^\text{query}_i; \theta'_i)$
        \EndFor
        \State Accumulate meta-gradient: $\nabla_\theta \sum_i \mathcal{L}_\text{task}(\mathcal{D}^\text{query}_i; \theta'_i)$
        \State Update meta-parameters: 
        \State $\theta \gets \theta - \beta \nabla_\theta \sum_i \mathcal{L}_\text{task}(\mathcal{D}^\text{query}_i; \theta'_i)$
    \EndWhile
    \State \textbf{return} $\theta$
\end{algorithmic}
\end{algorithm}

In Alg.\ref{alg:maml}, each $K$ correspond to one OD assignment, while each task $\mathcal{T}$ corresponds to a set of network closures over which $K$ ODs are assigned. This way, MAML trains the meta-parameters (namely, the meta-learned weights) over multiple different networks (each a different graph built over the same set of nodes). 

\begin{figure*}[!t]
    \centering
    \includegraphics[width=0.8\textwidth]{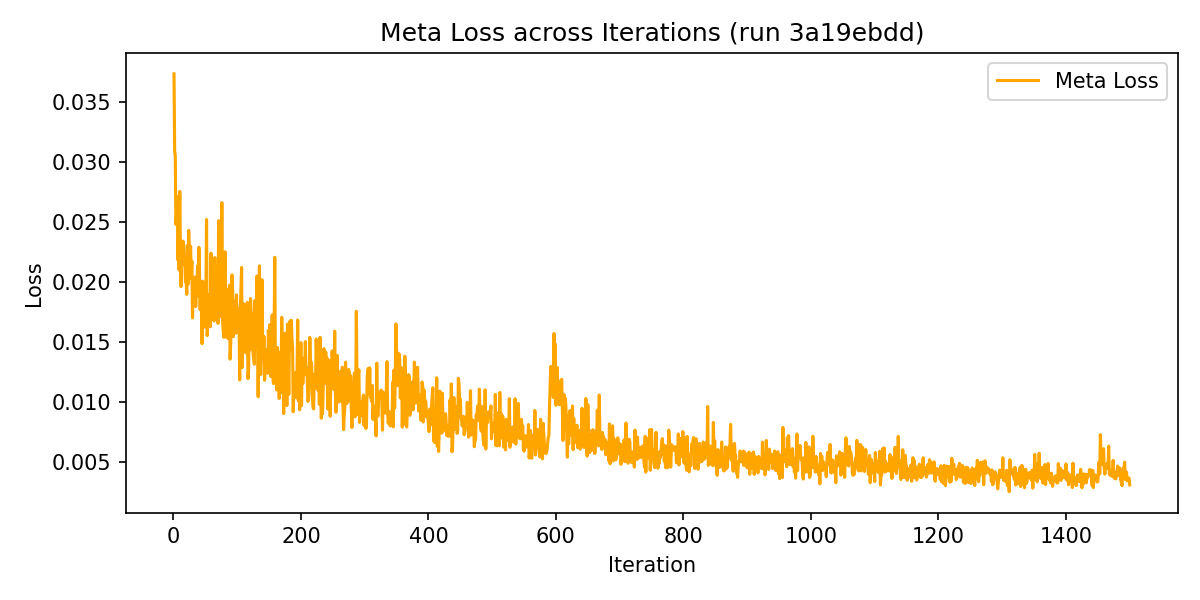}
    \caption{Meta-loss progression representing the improvements measured against the query set and in the outer loop, as the meta-training progresses}
    \label{fig:metaloss}
\end{figure*}

\section{The experimental setting}\label{sec:exp-set}

The described framework has been applied on a synthetic dataset built with the traffic assignment tool Aequilibrae \cite{b24}, an open source tool that performs static assignment through the bi-conjugate Frank-Wolfe algorithm. The case study of choice is the network of Eastern Massachussets, built from an open-source dataset \cite{b25} to maximize reproducibility. The road network is composed by 74 zones, 74 nodes and 258 edges. The base OD matrix includes 65576 trips. 

\subsection{The dataset}
To train the gatedGCN through the MAML architecture, we need to create a dataset of traffic assignments that is wide enough to cover different network closures, each with different OD matrices as input. For our experiments, we run 24864 Frank-Wolfe assignments, resulting in simulated flows for 336 different sets of closures, each with 74 different OD matrices. The 336 different versions of the network were defined by randomly deleting between 5\% and 30\% of the 258 edges. The different OD matrices have instead been generated by perturbing the base OD matrix following the methodology for spatial correlations described in \cite{b26}. The resulting OD distribution matches a symmetric Gaussian distribution with decreases/increases in trips ranging between 15\% and 70\%. 

\subsection{Support and query set}
The 24864 entries in the generated dataset were then separated in a support and a query set. As mentioned in Section~\ref{sec:methodology}, the former includes only a subset of OD entries for each task, to train the gatedGCN in an artificial condition of data scarcity. The latter instead includes a larger number of OD entries while imposing support $\notin$ query, and is used in the outer loop to "validate" the performance that the gatedGCN develops in the inner loop. In our experiments, at each iteration of the outer loop $i$ the MAML architecture draws support and query set so that there is no spill-over (i.e. support tasks $\notin$ query tasks) with the size of the support set equal to $K \cdot \mathcal{T}_i$. The query set's size is instead $M \cdot \mathcal{T}_i$. 

\subsection{Test set}

Once the MAML framework reaches a meta-loss in the outer loop below an acceptable threshold, the resulting meta-parameters are adopted as weights of the gatedGCN and tested in how they allow the gatedGCN to adapt once a new task (a new set of closures) is considered. It is paramount, to avoid overestimating the MAML performance, that these new tasks do not appear at any point during the meta-training (described in Alg.~\ref{alg:maml}). For this reason, at the start of the experiments, 3 network closures and 25 OD matrixes are extracted from the original dataset and do not appear in the support and query sets used for training. The test dataset is thus of size $3 \cdot 25 = 75$ assignments. As the proposed MAML framework aims at quick adaptation,  the test dataset is also split into support/query as the gatedGCN is allowed one round of inner loop steps on the support data (namely, the scarce OD data available to the modeler as the new set of closure arises). The resulting weights are then validated in their capacity to predict a wider set of OD scenarios on these hidden closures, i.e. the scenarios saved in the (test) query set.

\subsection{Hyperparameters}
In this subsection we report the adopted hyperparameters to ensure reproducibility. First, the hyperparameters reported in Table~\ref{tab:gcnhyperparameters} were used in the gatedGCN.

\begin{table}[h]
    \centering
    \begin{tabular}{|p{3cm}|p{2cm}|}
        \hline
         Epochs & 100 \\ \hline
         Batch size & 128 \\ \hline
         Hidden dimensions & 192 \\ \hline
         Layers & 6 \\ \hline
         Dropout & 0.01 \\ \hline
         Learning rate & 0.02 \\ \hline
    \end{tabular}
    \caption{Hyperparameters adopted for the gatedGCN}
    \label{tab:gcnhyperparameters}
\end{table}

Then, Table~\ref{tab:mamlhyperparameters} reports the hyperparameters relevant for the MAML architecture.

\begin{table}[h]
    \centering
    \begin{tabular}{|p{3cm}|p{2cm}|}
        \hline
         Meta-iterations & 1500 \\ \hline
         K & 4 \\ \hline
         M & 25 \\ \hline
         Inner loop steps & 5 \\ \hline
         $\mathcal{T}_i$ batch & 7 \\ \hline
         Learning rate & 0.055 \\ \hline
         
    \end{tabular}
    \caption{Hyperparameters adopted for the meta-training of the MAML architecture}
    \label{tab:mamlhyperparameters}
\end{table}

\section{Results}\label{sec:results}
In this section we report the results of the experiment. As Fig.~\ref{fig:metaloss} shows, the outer loss steadily decreases and plateaus around the value of 0.003. The trend proves that the MAML framework is successful in improving the initialization starting point, as each meta-iteration exploits the same amount of inner loop iterations (Alg.\ref{alg:maml}), the only difference being the starting weights themselves. 

The final weights, obtained for the lowest Loss value from Fig.~\ref{fig:metaloss}, are then applied to the 3 sets of network closures that have been kept hidden from the meta-training process. These are tasks 47, 248 and 293, randomly chosen out of the 336 different closure patterns. The meta-test results compare the performance of the gatedGCN across all the selected OD matrices for these closures, specifically how well the flows on each edge are predicted. The results are reported in Fig.~\ref{fig:net4}

\begin{figure}
    \centering
    \includegraphics[width=0.4\textwidth]{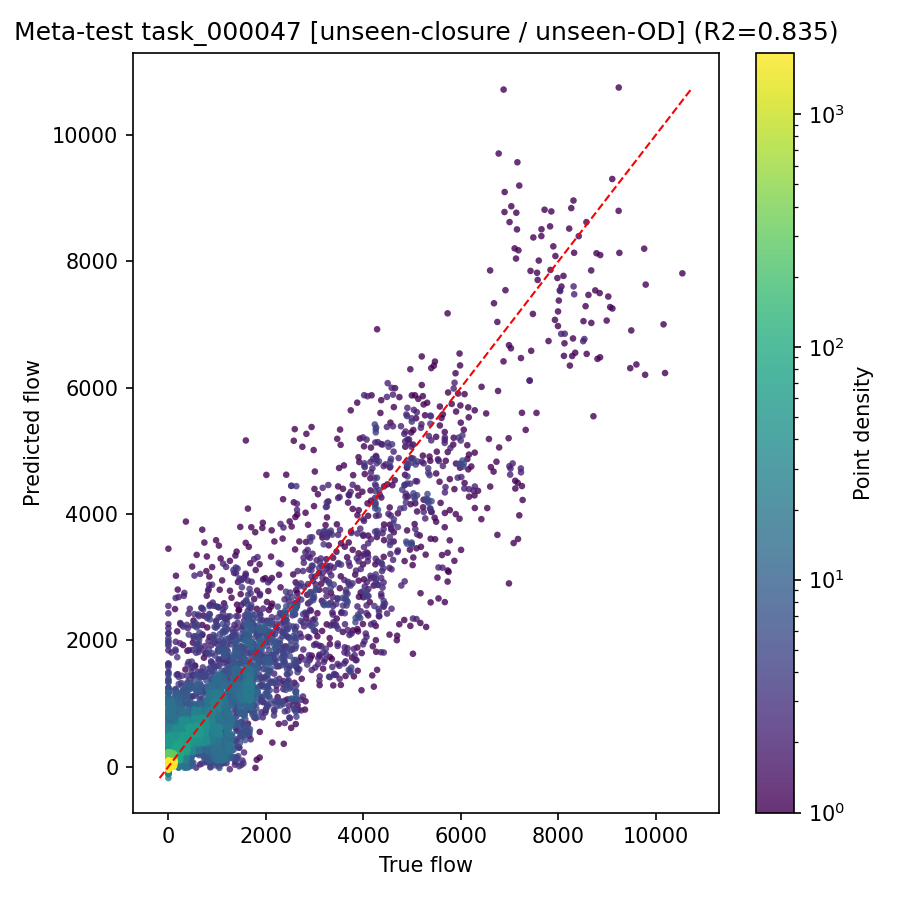}
    \includegraphics[width=0.4\textwidth]{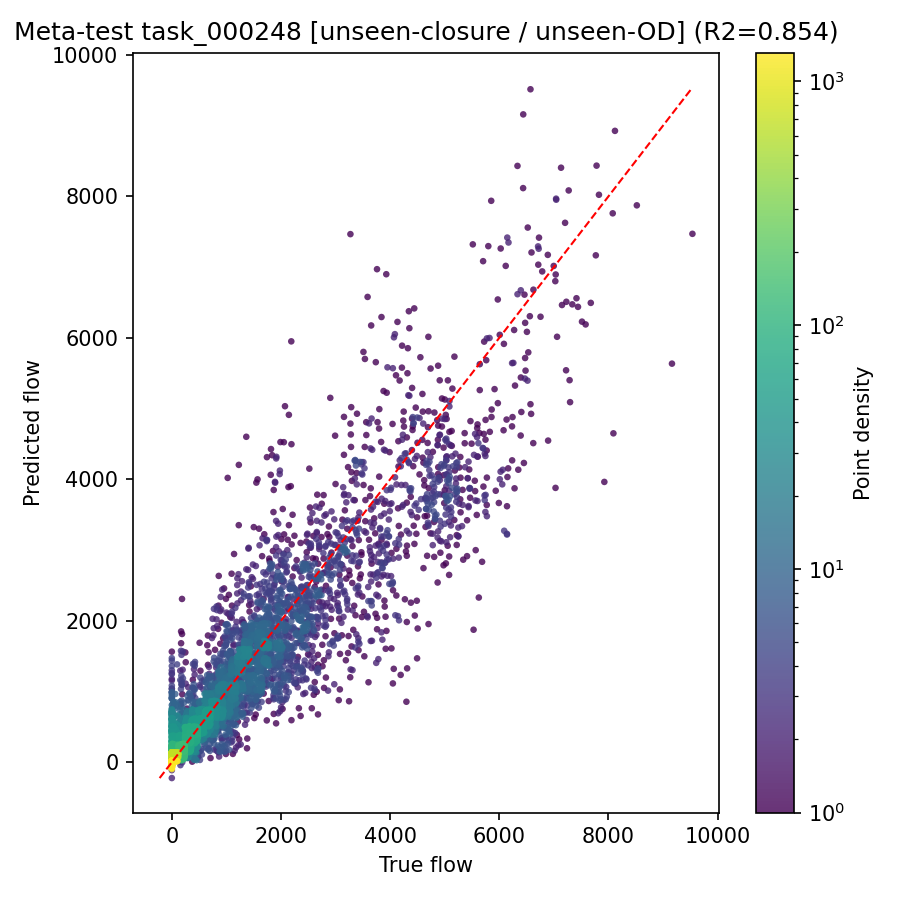}
    \includegraphics[width=0.4\textwidth]{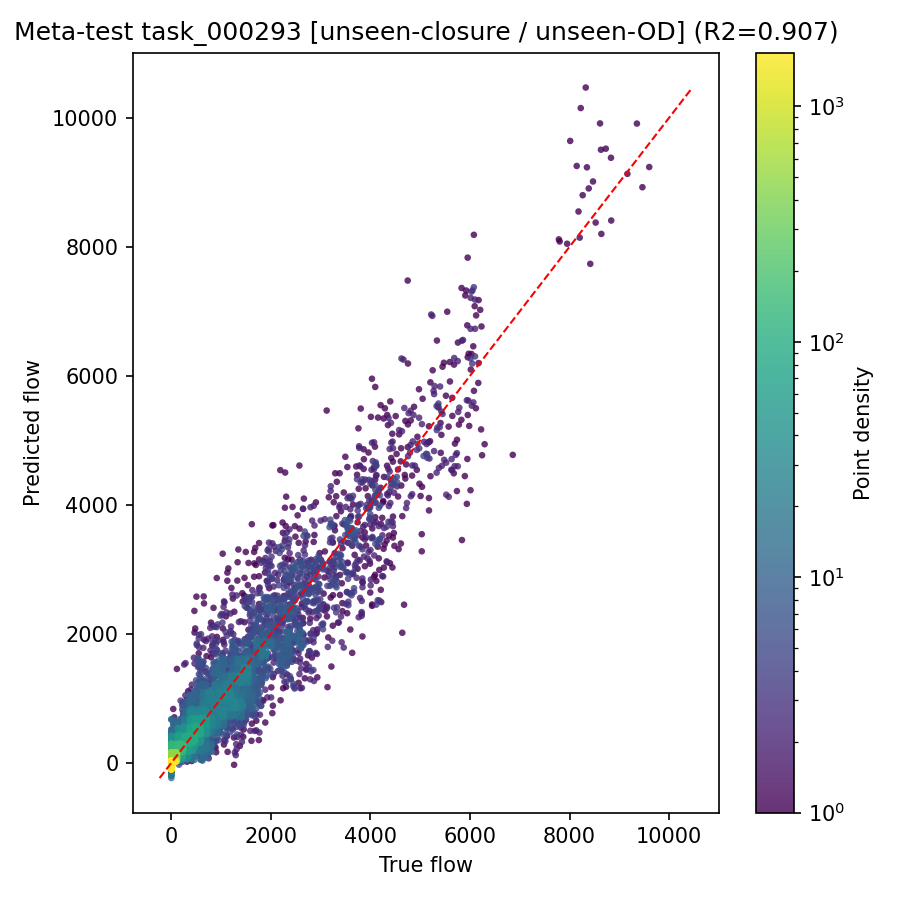}
    \caption{Scatter plots of predicted vs estimated flows. Each plot reports values for twice the number of links times the number of query OD matrices in the meta-test}
    \label{fig:net4}
\end{figure}

As it can be noticed, the initialized weights resulting from the meta-training are still able to provide solid estimates for network closures that were not seen during the meta-training itself, with new OD matrices. The setting is a particularly challenging one, as the gatedGCN not only must quickly adapt to new configurations but it must do so in a feature space that does not overlap with the training data (the unseen ODs). This is especially visible for $x = 0$ in the plots, as the MAML initialization allows the gatedGCN to correctly predict most of the closed edges (as shown by the density of points at (0,0)) but not all of them. The $R^2$ ranges (0.835
to 0.907) suggest that the closure pattern may play a relevant role in determining the capacity of the gatedGCN to quickly adapt, yet all the trends follow the 45 degree line that signals a consistent prediction, with isolated clusters of few data points as outliers. 

\subsection{Relevance to traffic best practices}
The reported experiments are meant to simulate real life use cases, in which stakeholders such as municipalities or road authorities may be faced with disruptive scenarios for which they have little data concerning both the demand and the supply. An example is the flooding scenario, in which different links across the network may be impacted, depending on the severity of the event \cite{b21}. Visually, these events may result in variations of the network such as the ones used for our experiments (Fig.~\ref{fig:networks}). As it can be seen, the MAML+GNN architecture has been tested in very challenging configurations where historic paths are not relevant for most OD pairs. It is unlikely that the stakeholders would have enough demand data in these cases to properly train the gatedGCN on this kind of disrupted graph, a theme that is especially relevant in medium-sized municipalities. By training the gatedGCN through our MAML framework, we allow it to maintain an acceptable level of performance without the need to retrain it with the new data and graph. This, in turn, allows for very fast response times, as faithful snapshots are available with the speed that characterizes state-of-the-art machine-learning solutions. 

\begin{figure}
    \centering
    \includegraphics[width=0.8\linewidth]{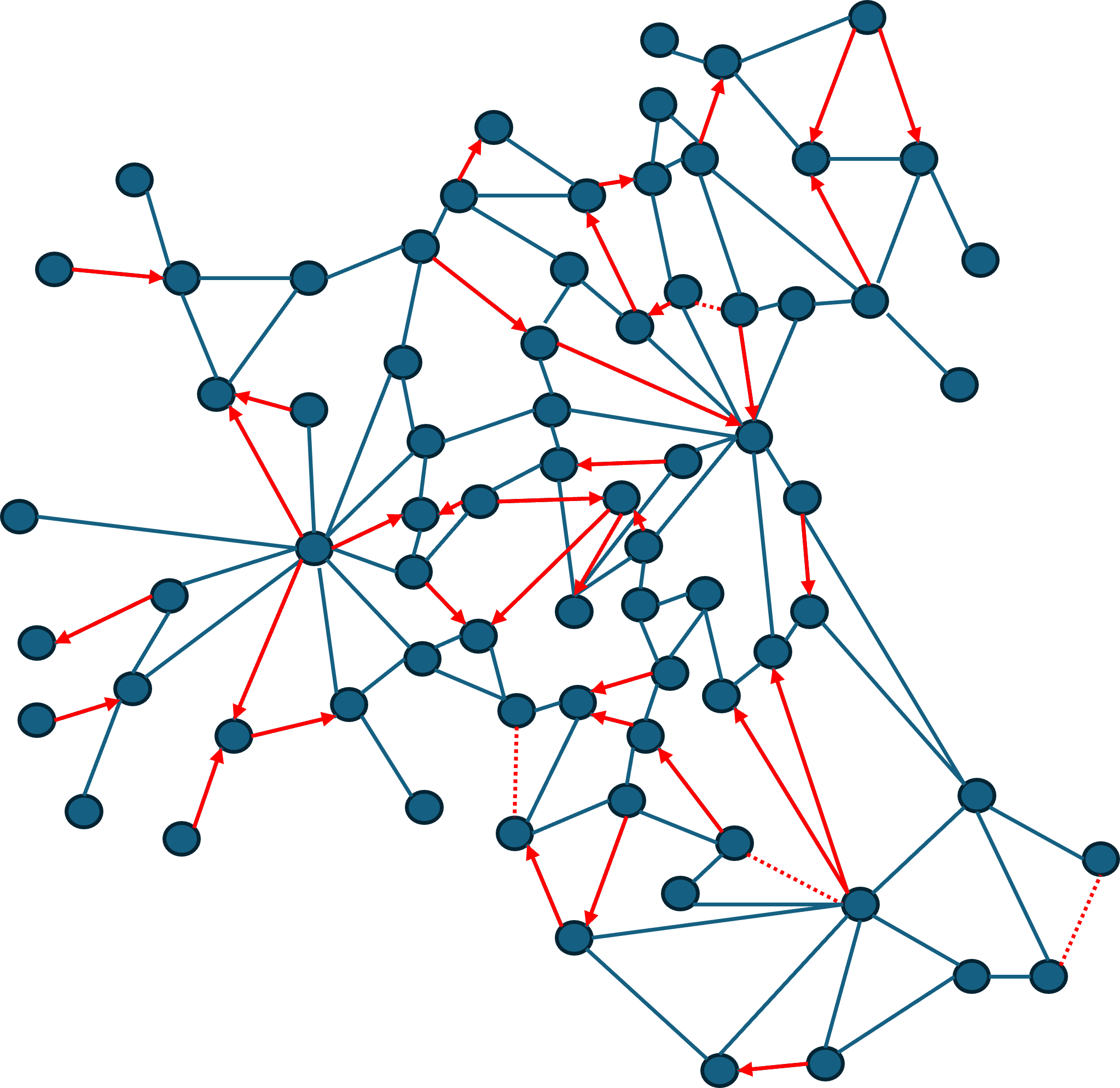}
    \caption{Network variation in task 47 - the red arrows signal the links that have become directed (one direction has been closed) while the dashed links represent closures in both directions.}
    \label{fig:networks}
\end{figure}

\subsection{Limitations and future research directions} 
The reported experiments provide a first application of the MAML framework on the problem of network disruptions, allowing the gatedGCN to learn the task of performing a traffic assignment on different versions of the network. Conceptually, our setting allows the learning of weights that quickly adapt to different graphs (and consequently road networks) as long as the nodes remain the same. A future research direction is to widen the MAML+GNN framework so that each task may use a completely different graph (i.e. varying edges \textit{and} nodes), which in turn would allow the gatedGCN to quickly adapt to network from different cities and not only on variations of the same network. 
Besides, the study reports initial results that focus on the task of predicting the assignment of flows across the network. While the OD demand is varied and split between meta-training and meta-test, there is currently no correlation between the network closures and the assigned demand matrix. The use of a more sophisticated simulation tool to generate the synthetic datasets would allow to build tasks where the demand adapts to the set of closures in the road network. 
Finally, the proposed framework may be applied to cities of different sizes and to different surrogates, to isolate the effect of different architectures on the MAML performance for traffic assignment tasks. 

\section{Conclusions}\label{sec:conclusions}
This study describes a variation of the MAML framework applied to initialize a gatedGCN for traffic assignment tasks. It differs from common approaches in literature as it does not focus on forecasting but allows instead to learn how traffic flows distribute on different graph structures, as long as nodes are unchanged. The results suggest that coupling MAML with a standard gatedGCN provides satisfactory results, despite the challenges arising from a changing graph. While the $R^2$ ranges between 0.8 and 0.9, the trends are consistently aligned with the 45 degree line of the scatter plots and the density of predicted/true flows suggest an high capacity of learning the effects of unseen closures on the network. The proposed framework can be applied for fast adaptation during disruptive events that modify the road network beyond patterns measured by historic data, allowing the use of a standard GNN model with no need for retraining.




\begin{thebibliography}{00}
\bibitem{b1} Lassen, O. B., Agriesti, S., Eldafrawi, M., Gammelli, D., Cantelmo, G., Gentile, G., and Pereira, F. C. Learning Traffic Flows: Graph Neural Networks for Metamodelling Traffic Assignment. In 2025 9th International Conference on Models and Technologies for Intelligent Transportation Systems (MT-ITS) (pp. 1-8). IEEE.
\bibitem{b2} Rahman, R., and Hasan, S. Data-driven traffic assignment: A novel approach for learning traffic flow patterns using graph convolutional neural network. Data Science for Transportation, 5(2), pp. 11, 2023.
\bibitem{b3} Veres, M., and Moussa, M. Deep learning for intelligent transportation systems: A survey of emerging trends. IEEE Transactions on Intelligent transportation systems, 21(8), pp. 3152–3168, 2019.
\bibitem{b4} Ye, Y., Xiao, Y., Zhou, Y., Li, S., Zang, Y., and Zhang, Y. Dynamic multi-graph neural network for traffic flow prediction incorporating traffic accidents. Expert Systems with Applications, 234, 121101, 2023.
\bibitem{b5} Waikhom, L., Patgiri, R., and Singh, L. D. Dynamic temporal position observant graph neural network for traffic forecasting: Waikhom et al. Applied Intelligence, 53(20), pp. 23166-23178, 2023.
\bibitem{b6} Chen, J., Zheng, L., Hu, Y., Wang, W., Zhang, H., and Hu, X. Traffic flow matrix-based graph neural network with attention mechanism for traffic flow prediction. Information Fusion, 104, pp. 102146, 2024.
\bibitem{b7} Huang, X., Ye, Y., Yang, X., and Xiong, L. Multi-view dynamic graph convolution neural network for traffic flow prediction. Expert Systems with Applications, 222, pp. 119779, 2023.
\bibitem{b8} Khan, R. H., Miah, J., Arafat, S. Y., Syeed, M. M., and Ca, D. M. Improving traffic density forecasting in intelligent transportation systems using gated graph neural networks. In 2023 15th international conference on innovations in information technology (IIT) (pp. 104-109), IEEE, 2023.
\bibitem{b9} Gammelli, D., Yang, K., Harrison, J., Rodrigues, F., Pereira, F. C., and Pavone, M. Graph neural network reinforcement learning for autonomous mobility-on-demand systems. In 2021 60th IEEE Conference on Decision and Control (CDC) (pp. 2996-3003), IEEE, 2021.
\bibitem{b10} Ouyang, X., Yang, Y., Zhang, Y., Zhou, W., Wan, J., and Du, S. Domain adversarial graph neural network with cross-city graph structure learning for traffic prediction. Knowledge-Based Systems, 278, pp. 110885, 2023.
\bibitem{b11} Rampasek, L.,  Galkin, M., Dwivedi, V., Luu, A., Wolf, G., and Beaini, D. Recipe for a General, Powerful, Scalable Graph Transformer. Advances in Neural Information
Processing Systems, 35, 2022.
\bibitem{b12} Liu, T., and Meidani, H. End-to-end heterogeneous graph neural networks for traffic assignment. Transportation Research Part C: Emerging Technologies, 165, pp. 104695, 2024.
\bibitem{b13} Liu, T., and Meidani, H. Multi-class traffic assignment using multi-view heterogeneous graph attention networks. Expert Systems with Applications, 286, pp. 128072, 2025.
\bibitem{b14} Hu, X., and Xie, C. Use of graph attention networks for traffic assignment in a large number of network scenarios. Transportation Research Part C: Emerging Technologies, 171, pp. 104997, 2025
\bibitem{bmeta} Finn, C., Abbeel, P., and Levine, S. Model-agnostic meta-learning for fast adaptation of deep networks. In International conference on machine learning (pp. 1126-1135), PMLR, 2017.
\bibitem{b16} Roark, A., Agriesti, S., Pereira, F. C., and Cantelmo, G. Learning to Learn the Macroscopic Fundamental Diagram using Physics-Informed and meta Machine Learning techniques. arXiv preprint arXiv:2508.14137, 2025.
\bibitem{b17} Wang, C., Zhao, F., Luo, H., Sun, P. Z., and Fang, Y. Cross-transportation-mode knowledge transfer for trajectory recovery with meta learning. IEEE Transactions on Intelligent Transportation Systems, 2025.
\bibitem{b18} Jiang, Z., Huang, A., Jiang, R., Chen, J., Sekimoto, Y., and Guan, W. Leveraging Spatial–Temporal Heterogeneity and Cross-Mode Interactions: A Meta-Learning Approach for Multimodal Transportation Demand Prediction. IEEE Transactions on Intelligent Transportation Systems, 2025.
\bibitem{b19} Fan, Y., Xu, J., Zhou, R., and Liu, C. Transportation-mode aware travel time estimation via meta-learning. In International Conference on Database Systems for Advanced Applications (pp. 472-488), Cham: Springer International Publishing, 2022.
\bibitem{b20} Liu, G., Zhang, X., Qian, Z., Chen, L., and Bi, Y. Life cycle assessment of road network infrastructure maintenance phase while considering traffic operation and environmental impact. Journal of Cleaner Production, 422, pp. 138607, 2023.
\bibitem{b21} Costa, M., Petersen, M. W., Vandervoort, A., Drews, M., Morrissey, K., and Pereira, F. C. Climate Adaptation with Reinforcement Learning: Experiments with Flooding and Transportation in Copenhagen. https://arxiv.org/abs/2409.18574, 2024.
\bibitem{b22} Bresson, X., and Laurent, T. Residual Gated Graph ConvNets. https://arxiv.org/abs/1711.07553, 2018
\bibitem{b23} PyTorch. SmoothL1Loss [Online:] https://docs.pytorch.org/docs/stable/ generated/torch.nn.SmoothL1Loss.html\#torch.nn.SmoothL1Loss
\bibitem{b24} Camargo, P. AequilibraE and Tradesman: current state of affairs of modelling with open-source software. In Australasian Transport Research Forum (ATRF), 44th, 2023.
\bibitem{b25} Transportation Networks for Research Core Team. Transportation Networks for Research. [Online:] https://github.com/bstabler/TransportationNetworks. 
\bibitem{b26} Qurashi, M., Lu, Q., Cantelmo, G., and Antoniou, C. PC-SPSA: Exploration and assessment of different historical data--set generation methods for enhanced DTA model calibration. In 9th Symposium of the European Association for Research in Transportation (hEART2020), 2021.

\end{thebibliography}
\end{document}